\documentclass{article}
\usepackage{natbib}
\usepackage{minipage}
\usepackage{tcolorbox}
\usepackage{ulem}
\usepackage{wrapfig}

\usepackage{multirow}
\usepackage{array}
\usepackage{tabularx}
\usepackage{makecell}
\usepackage{colortbl}
\usepackage{CJKutf8}
\usepackage{graphicx}
\usepackage{enumitem}

\usepackage[final]{neurips_2025}
\usepackage[utf8]{inputenc} % allow utf-8 input
\usepackage[T1]{fontenc}    % use 8-bit T1 fonts
\usepackage{hyperref}       % hyperlinks
\usepackage{url}            % simple URL typesetting
\usepackage{booktabs}       % professional-quality tables
\usepackage{amsfonts}       % blackboard math symbols

\usepackage{microtype}      % microtypography
\usepackage{xcolor}         % colors

\definecolor{lightgray}{gray}{0.9}
\definecolor{headergreen}{RGB}{92, 184, 92}

\definecolor{positive}{HTML}{0072B2} % Define color using hex code
\definecolor{neutral}{HTML}{666666} % Define color using hex code
\definecolor{baseline}{HTML}{666666} % Define color using hex code
\definecolor{negative}{HTML}{D55E00} % Define color using hex code

\usepackage{color-edits}
\addauthor{gn}{magenta}
\addauthor{sw}{cyan}

\title{Checklists Are Better Than Reward Models For Aligning Language Models}

\author{\textbf{Vijay Viswanathan }$^{\heartsuit}$\ \ \ \  
Yanchao Sun$^{\clubsuit}$\ \ \ 
Shuang Ma$^{\clubsuit}$\thanks{Work performed while at Apple.}\ \ \ \ 
Xiang Kong$^{\clubsuit}$\ \ \ \  \\  
\textbf{Meng Cao}$^{\clubsuit}$\ \ \ \  
\textbf{Graham Neubig}$^{\heartsuit}$\ \ \ \  
\textbf{Tongshuang Wu}$^{\heartsuit}$\\
$^{\heartsuit}$ Carnegie Mellon University\ \ \  $^{\clubsuit}$Apple
}

\begin{document}

\maketitle

\setcounter{footnote}{0}

\begin{abstract}
% un-formatted version:
% Language models must be adapted to understand and follow user instructions. Reinforcement learning is widely used to facilitate this— typically using fixed criteria such as "helpfulness" and "harmfulness". In our work, we instead propose using flexible, instruction-specific criteria as a means of broadening the impact that reinforcement learning can have in eliciting instruction following. We propose "Reinforcement Learning from Checklist Feedback" (RLCF). From instructions, we extract checklists and evaluate how well responses satisfy each item—using both AI judges and specialized verifier programs—then combine these scores to compute rewards for RL. We compare RLCF with other alignment methods applied to a state-of-the-art instruction following model (Qwen2.5-7B-Instruct) — RLCF is the only method to improve on every benchmark, including a 4 point increase in hard satisfaction rate on FollowBench and a 3 point boost in win rate on Arena-Hard. These results establish checklist feedback as a key tool for improving language models' support of queries that express a multitude of needs. We will release our models and our dataset of checklists, "WildChecklists", to the public.
Language models must be adapted to understand and follow user instructions. Reinforcement learning is widely used to facilitate this -- typically using fixed criteria such as ``helpfulness'' and ``harmfulness''. In our work, we instead propose using flexible, instruction-specific criteria as a means of broadening the impact that reinforcement learning can have in eliciting instruction following.
%\gncomment{be consistent in hyphenation of finetuning}
We propose ``\textbf{Reinforcement Learning from Checklist Feedback}'' (\textbf{RLCF}). From instructions, we extract checklists and evaluate how well responses satisfy each item—using both AI judges and specialized verifier programs—then combine these scores to compute rewards for RL. We compare RLCF with other alignment methods on top of a strong instruction following model (\texttt{Qwen2.5-7B-Instruct}) on five widely-studied benchmarks -- \textbf{RLCF is the only method to help on every benchmark}, including a 4-point boost in hard satisfaction rate on FollowBench, a 6-point increase on InFoBench, and a 3-point rise in win rate on Arena-Hard. We show that RLCF can also be used off-policy to improve \texttt{Llama 3.1 8B Instruct} and \texttt{OLMo 2 7B Instruct}.
These results establish rubrics as a key tool for improving language models' support of queries that express a multitude of needs. We release our our dataset of rubrics (\textit{WildChecklists}), models, and code to the public.\footnote{Code: \url{www.github.com/viswavi/RLCF}, Dataset: \url{www.huggingface.co/datasets/viswavi/rlcf}}
\end{abstract}

\definecolor{lightgray}{gray}{0.9}
\definecolor{headergreen}{RGB}{92, 184, 92}

\section{Introduction}

Language models must follow user instructions to be useful. As the general public integrates language model-based assistants into their completion of daily tasks, there is an expectation that models can faithfully follow the users' requests, which increasingly involve rich, multi-step instructions \citep{Liu2024WeNS, Zhao2024WildChat1C, zhenglmsys}.
%For example, consider a user request to "Write a Python script that, given a set of documents, produces 20 clusters using word frequency statistics and writes the output as a CSV file." Here, the user is asking an assistant to simultaneously handle (1) clustering algorithm selection and execution, (2) feature extraction, and (3) output formatting in a single response.
Today's models are almost universally trained to follow instructions via a two-step process: instruction finetuning, followed by reinforcement learning from human feedback (RLHF). Instruction finetuning, where the model is trained to mimic responses generated by annotators \citep{Raffel2019ExploringTL}, has historically been the primary workhorse for imbuing language models with some amount of instruction following ability \citep{Wang2022SelfInstructAL, Chung2022ScalingIL, Xu2024MagpieAD, Lambert2024TLU3P}. Model developers then frequently employ RLHF, where the model is trained to generate responses that look more like labeled ``good'' responses than ``bad'' responses, as a refinement step to decrease the likelihood that the model exhibits predefined poor behaviors (typically harmful behaviors) \citep{Ziegler2019FineTuningLM, Bai2022ConstitutionalAH}. Unlike ``verifiable'' tasks where reinforcement learning is a workhorse \citep{DeepSeekAI2025DeepSeekR1IR, Lambert2024TLU3P, ifbench}, reinforcement learning remains difficult to utilize for ambiguous or ``non-verifiable'' tasks, such as instruction following. What would it take to make RL a general-purpose solution at eliciting desirable behaviors in subjective or open-ended settings?
% \swcomment{should we provide some reasons for why we think RL should be utilized more? e.g. just instruction tuning is not enough because they might have limited generalizability / is sparse, and RL should be more flexible, so we shouldn't constraint it to just decrease undesirable behaviors but use it as a core step for instruction following?}

%finetuning on SFT data can still yield instruction following this data is either small or noisy, but the performance declines \citep{Hewitt2024InstructionFW}.
% \swcomment{Is this paragraph about: RL is not used to its full potential because we are not getting the right signals? If so, maybe replace "...Each of these approaches has a limitation" and simply just start with this statement? And then we can describe these three approaches and give a summary on all of them --- they either compare responses in a too sparse way (which leads to the problem of reward hacking and XXX), or they have limited coverage on what dimensions could be checked (needs to be verifiable, or need to be carefully crafted beforehand and therefore limits to e.g. safety alignments etc.?) This might help better tie this paragraph with the next. Right now it's not very easy to decipher that these limitations all lead to a need of "easier to automatically verify at scale"}

\setlength{\belowcaptionskip}{-20pt}
\begin{wrapfigure}[15]{r}{0.44\textwidth}
  \centering
  \vspace{-10pt}
\includegraphics[width=0.99\linewidth]{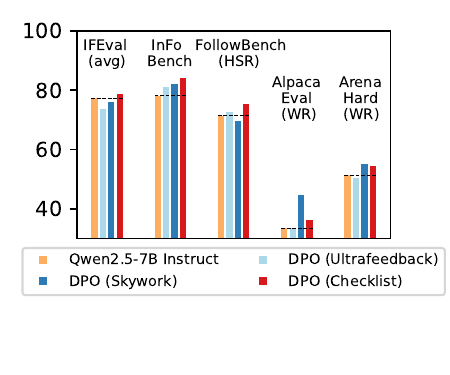}
  \vspace{-44pt}
  \caption{RL on Checklist Feedback consistently improves Qwen2.5-7B-Instruct, whereas every other source of automatic feedback gives mixed results.}
  \label{fig:teaser_figure}
\end{wrapfigure}

We believe the solution must involve producing better reward signals.
Recent work on RL for language model alignment has focused on automatically obtaining feedback on model behavior, either by (1) exclusively using \textit{instructions with verifiable answers} \citep{autoif, ifbench}, (2) grading responses with specially-trained reward models \citep{Wang2024HelpSteer2PreferenceCR, Eisenstein2023HelpingOH}, or (3) distilling preferences from a larger prompted model \citep{Bai2022ConstitutionalAH, Tunstall2023ZephyrDD}. Using instructions with verifiable answers can only reinforce limited aspects of behavior (ignoring subjective constructs, e.g. topicality or style). Reward models (``RMs'') are flexible, but their notion of rewards can be arbitrary, leading to reward hacking \citep{Eisenstein2023HelpingOH}. When distilling preferences from a larger model, that model must infer what aspects to grade on, reducing the ``generator-verifier gap'' that enables RL \citep{Swamy2025AllRL}. Even if multiple criterion-specific prompts are used, these criteria may not be comprehensive \citep{Bai2022ConstitutionalAH, Glaese2022ImprovingAO}. 
% \gncomment{Should we contrast with rule-conditioned reward or constitutional AI, which have global checklists \citep{glaese2022improving,Bai2022ConstitutionalAH}}

In this paper, we ask: ``how can we grade responses to instructions in a manner that is \textit{automatic} (requires no human annotation), \textit{flexible} (considers all aspects of response quality), \textit{intuitive} (aligned with perceptible differences in responses), and \textit{applicable to any instruction or response}, to enable more better language model alignment?''
%In this paper, we ask: ``how can we make instruction following easier to automatically verify at scale''?
%\swcomment{The RQ seems a bit broad -- Should we try a more specific version like "how can we make more faithful and generalizable assessments on model responses, such that we can enable more effective use of RL for instruction following training?"}
%We more specifically frame this question as: ``how can we grade responses to instructions in a manner that is \textit{automatic}, \textit{clearly aligned with perceptible aspects}, \textit{flexible}, and \textit{applicable to almost any instruction-response pair}''?
We propose extracting dynamic rubrics from instructions -- an approach we term \textbf{Reinforcement Learning from Checklist Feedback (RLCF)}. 
%\swcomment{"parsing any instructions into custom/dynamic checklists"}
This approach reduces the task of grading responses to answering a series of yes/no questions, which can be answered by a model or by executing a verification program.

\begin{figure}[t]
    
  \centering
\includegraphics[trim={0 21.5cm 21.8cm 0cm}, clip, width=0.95\linewidth]{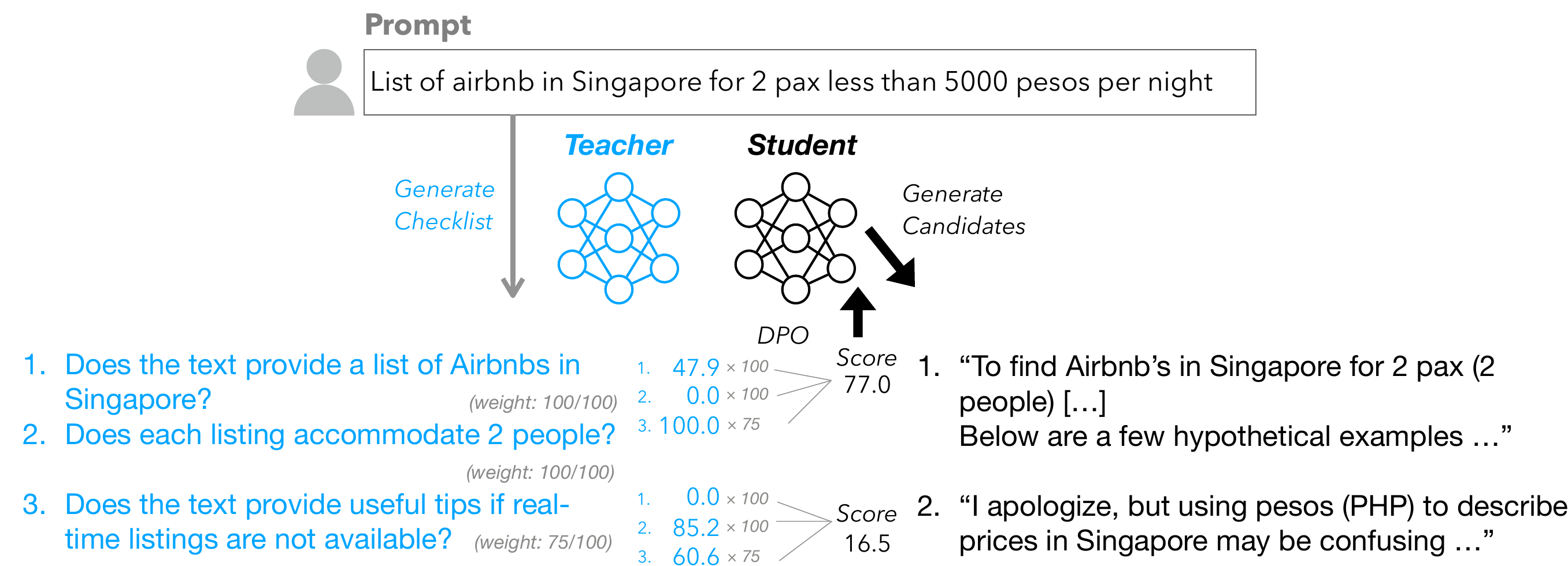}
  \caption{We propose \textit{Reinforcement Learning from Checklist Feedback}, where sampled responses are evaluated by a teacher model grounded on a fixed set of criteria. In our pipeline, given instructions, we first generate checklists synthetically from the instructions, grade each response on each checklist item, combine per-item scores into a single weighted checklist score, then use this score for RL. }
  \label{fig:main_figure}
\end{figure}

%\swcomment{I think I'm still not 100\% clear whether our target baseline is instruction-finetuning or other reinforcement learning methods. Most of the existing intro seems to talk about instruction tuning, but feels like our actual baseline is RL. So, should we be talking more about limitations of RL methods (and after talking about numerical gains also explain qualitatively what's the strength of checklist-rewards), and less about the instruction finetuning / availability of instruction dataset?}

Our key contributions are:
\begin{enumerate}[leftmargin=2em, itemsep=1pt, parsep=0pt, topsep=1pt]
    \item  We describe a new and improved algorithm for automatically generating checklists at scale. 
    \item We construct \textit{WildChecklists}, a dataset consisting of 130,000 instructions and corresponding checklists (generated synthetically). When applicable, we accompany items in each checklist with a verification program to facilitate automatic evaluation. We plan to release this dataset to the community as an artifact for future study.
    \item We describe a new algorithm for grading responses according to checklists, using language models and code, and we show to use this algorithm to rank responses for preference tuning.
    \item We finetune \texttt{Qwen2.5-7B-Instruct} via reinforcement learning from checklist feedback using \textit{WildChecklists}, leading to a strong and improved 7B-parameter model for instruction following.
\end{enumerate}

On 5 benchmarks covering both constrained instruction following (IFEval, InFoBench, FollowBench) and general conversational assistance (AlpacaEval, Arena-Hard), we find that RLCF provides benefits on all instruction following benchmarks while maintaining improved performance on general conversational assistance benchmarks. In contrast, all alternative forms of AI feedback lead to mixed results, as shown in \autoref{fig:teaser_figure}. RLCF provides a  5.4\% relative improvement over \texttt{Qwen2.5-7B-Instruct} in average hard satisfaction rate on FollowBench, a 6.9\% relative improvement in overall requirement following ratio on InFoBench, and a 6.4\% relative improvement on Arena-Hard \citep{Jiang2023FollowBenchAM, qin2024infobench, ArenaHard}. RLCF can also be used off-policy; we see \texttt{Llama 3.1 8B Instruct} and \texttt{OLMo 2 7B Instruct} improve using samples collected using \texttt{Qwen2.5-7B-Instruct}. Despite these considerable improvements, RLCF simply requires a teacher model, with no need for additional data or human annotations, making this approach amenable to diverse languages or domains.
We provide evidence that checklist-based rewards are well-correlated to human preference judgments (comparable to many finetuned reward models) while providing a stronger learning signal than alternatives.
% \swcomment{add some more numbers to this result paragraph, and maybe some speculation on why RLCF work better?}

\section{Checklist Generation}
\label{sec:checklist_generation}
%\swcomment{before jumping into the method, should we make some definitions on e.g. what's a checklist, what additional information checklists provide, etc.?}
\textbf{Desiderata for checklists.}
We define a \textit{checklist} as a sequence of \textit{requirements} paired with an instruction that satisfy the following properties:
\begin{enumerate}[leftmargin=2em, itemsep=1pt, parsep=0pt, topsep=1pt]
    \item Each requirement in the checklist is a yes/no question (e.g. ``Does the text contain 3 commas?'').
    \item Each requirement in the checklist must be answered relative to a given candidate response.
    \item A response would be considered acceptable if and only if the response answers ``yes'' to all checklist requirements.
\end{enumerate}

To satisfy definition \#3, checklists must be \textit{comprehensive} (cover most relevant aspects of quality) and \textit{natural} (entailed by their corresponding instructions). Based on the observation that false positive rewards are often more detrimental to reinforcement learning than false negatives \citep{Huang2024TheDS}, we want checklists that are \textit{objective} (facilitate automatic verification) and \textit{atomic} (each requirement focuses on a single aspect of quality), to make requirement checking easier.

\textbf{Extract checklists per instruction.} We examine two methods to extract checklists:

% \gncomment{I added titles to each paragraph. You named them ``Direct'' and ``With Candidates'' in the table, but what about ``Candidate-based'', it seems a little bit more like a method name.}

\begin{itemize}[leftmargin=2em, itemsep=1pt, parsep=0pt, topsep=0pt]
\item \texttt{Direct:} We simply prompt an LM to extract a checklist from a given instruction \citep{Cook2024TICKingAT}. This approach is intuitive and simple but risks repeating the original instruction via these individual criteria, which may limit \textit{comprehensiveness} and \textit{objectiveness}.

\item \texttt{Candidate-based:}
We view a requirement as any aspect of an instruction that, when absent, causes a response to fail. We propose a two-stage approach: produce responses of varying quality, then prompt an LM to write a checklist of all their possible failure modes. For each checklist item, we also prompt the model to generate an ``importance'' weight (from 0 to 100).
\end{itemize}

To compare these, we generate checklists for all instructions in InFoBench \citep{qin2024infobench}. We use \texttt{gpt-4o} to blindly evaluate each of these checklists on naturalness, objectivity, comprehensiveness, and atomicity, then select the better one overall. 
%With the source of each checklist concealed, we use \texttt{gpt-4o} to annotate whether these checklists are \textit{natural}, \textit{objective}, \textit{comprehensive}, or \textit{atomic}, before finally annotating whether which of the two checklists is better overall.  with the source of the checklist concealed. 
We manually perform the same evaluation on a \textit{subset of 50 instructions} from the ``Easy Set'' of InFoBench.
%We then perform the same annotation manually a subset of 50 instructions from the 
% \swcomment{do we have human-model agreement?}

The results in \autoref{tab:req_gen} show that checklists generated by prompting an LLM directly are more natural. However, providing candidate responses to the LLM leads to checklists with consistently better objectiveness, atomicity, and overall quality. There are absolute differences between scores from the two evaluations -- partly because they use different subsets -- but directional trends are consistent. We find that this difference translates to downstream performance after performing RL training. In \autoref{sec:comparing_req_gen_methods}, we show that Reinforcement Learning from Checklist Feedback is more effective on checklists generated via the candidate-based method.

\textbf{Regularization via universal criteria.}
In initial experiments, we found that optimizing for checklist completion led to responses beginning with long preamble overviews, suggesting reward hacking. Following \citet{Sun2023SALMONSW} (who report a similar issue in prior work), we add two ``universal requirements'' to all generated checklists. These requirements state ``\textit{1) The response directly address the request without excessive or off-topic information not necessary for addressing the user's instruction?} and \textit{2) The response should match the context and the instruction, whether it requires professionalism, friendliness, formality, or neutrality.}'', with a corresponding total importance weight of 100/100.
% \swcomment{should we argue why we design these universal requirements?}

\begin{table}[t]
\centering
\renewcommand{\arraystretch}{0.95}

\begin{tabular}{r |  c c | c  c}
\toprule
& \multicolumn{2}{c}{Manual Evaluation} & \multicolumn{2}{c}{Automatic Evaluation} 
\\
 \textit{\textbf{Metric}} &  \textbf{Direct} &  \textbf{Candidate-Based} &  \textbf{Direct} &  \textbf{Candidate-Based}  \\
\midrule
Naturalness &\textbf{ 94.9} & 93.9 & \textbf{88.0} & 85.1  \\
Objectiveness & 88.5 & \textbf{91.9} & 88.9 & \textbf{89.7} \\
Comprehensiveness & 74.0 & \textbf{82.0} & \textbf{69.2} & 64.8  \\
Atomicity & 68.0 & \textbf{90.0} & 98.6 & \textbf{99.0}  \\
\midrule
\% Preferred Overall & 38.0 & \textbf{56.0} & 40.6 & \textbf{51.2} \\
\bottomrule
\end{tabular}
\vspace{3pt}
\caption{We evaluate two checklist generation methods on four specific aspects of quality and an overall preference. Manual evaluation is performed on the first 50 rows of InFoBench ``easy'', while automatic evaluation is performed by \texttt{gpt-4o} on all 500 rows of InFoBench.}
\label{tab:req_gen}
\vspace{-15pt}
\end{table}

% \swcomment{This reads like we shouldnt trust the evaluation result...?}
\textbf{Dataset Generation} Using the candidate-based method, we generate checklists
%\gncomment{Which one is your proposed method?}
for 130,000 instructions from WildChat to create a new dataset, \textit{WildChecklists}. To generate candidate responses for our method, we use \verb+Qwen2.5-0.5B+, \verb+Qwen2.5-1.5B+, \verb+Qwen2.5-3B+, and \verb+Qwen2.5-7B+ \citep{Yang2024Qwen25TR}. \verb+Qwen2.5-72B-Instruct+ is the checklist generator model for both methods.

% \swcomment{In general, I feel sec 2 is a bit like a step-by-step description of our method right now. Might be good to at least split them by purpose, e.g. Define checklist, Generate checklist (2.1 and 2.2), and verify checklist (llm as a judge and code - 2.3 + 3.2?).} \gncomment{+1}

\section{Reinforcement Learning from Checklist Feedback}
\label{method}

Given \textit{WildChecklists}, we generate high-quality preference data for RL via a four-step process:

%\swcomment{space consideration: since each subsec is a single paragraph I think we can just use paragraph or textbf macro...}
\textbf{Sampling Candidate Responses.}
\label{sec:sampling}
To facilitate RL, we sample response pairs from our base policy with a temperature of 1.3 and a \textit{top-p} of 0.9 \citep{Holtzman2019TheCC}. This is simpler than prior works that systematically perturb samples to induce greater complexity \citep{conifer, autoif}.
%-- such synthetic modifications leads to an unnatural prompt distribution and leads to a more complicated method \gncomment{This sentence doesn't seem that important. At the most it should be a footnote, but it could also just be removed.

\textbf{Flexible Scoring}
\label{sec:scoring_with_judge}
Given a prompt, a response, and an individual checklist item, we use a combination of an LM judge and a verifier program to grade the response. For each checklist item, we prompt a judge model (\texttt{Qwen2.5-72B-Instruct}) using the prompt in \autoref{sec:generate_verification_code} to generate a numerical score between 0 and 100. We take the average of 25 numerical scores sampled from the model.\footnote{We sample responses using the \texttt{n} parameter in vLLM \citep{Kwon2023EfficientMM}. This approach follows prior work that describes the importance of using the mean score rather than mode score from an LM-as-a-judge model \citep{Wang2025ImprovingLI}
Regardless, this makes the AI judge component the computational bottleneck of our pipeline. In \autoref{sec:num_judges}, we show that \texttt{n} can be significantly reduced, at a modest accuracy cost. }.

LLMs struggle to evaluate hard, discrete criteria, such as ``does the response contain the letter R at least three times?''    \citep{Fu2024WhyDL}. To handle such constraints, we follow prior work in generating a program to grade responses when needed \citep{autoif, Zhou2023InstructionFollowingEF}. We prompt a model (using the prompt in \autoref{sec:generate_verification_code}) to write code only when the model is confident it can exactly check the requirement. For example, our program generator abstains from writing a verifier for the criterion ``Is the sentence coherent''. The program-graded score is then averaged with the AI judge's score.\footnote{This approach is much simpler than the most relevant prior work that uses programs to evaluate responses, AutoIF \citep{autoif}, which uses test-case generation and LM-based filters to remove low-quality programs.}
% \swcomment{is this more for appendix?}
%\swcomment{should we combine the above two paragraph to something like "adaptive scoring"?}

\textbf{Preference Tuning.}
Given a separate numerical score for each criterion for each response, we take the average of these scores, weighted by the importance score of each criterion. We keep only the 40\% of response pairs with the greatest difference along at least one criterion of its corresponding checklist. This removes response pairs that are too similar to offer useful pairwise signal.
%For example, for Qwen2.5-7B-Instruct, this means we keep the prompts and responses where the two responses differ along at least one criterion by 30 points (e.g. on the criterion of ``is the response written politely'', one response achieves a score of 75 and the other achieves a scores of 30 \gncomment{This example is super-specific and possibly a bit confusing, maybe it could just be deleted?}).
%\swcomment{30.1 feels like a very specific number lol How is ``.1'' important?}
We then assign these responses as a preference pair for direct preference optimization \citep{Rafailov2023DirectPO}.

\section{Experimental Setup and Results}
\label{sec:experimental_setup}

\subsection{Experimental Details}
\label{sec:experimental_details}
\textbf{Training Data} As a fixed source of instructions for all methods, we use WildChat, a set of natural conversations between users and AI language models crowdsourced from users across the world \citep{Zhao2024WildChat1C}. We filter out conversations that are non-English, toxic, or longer than two turns.

\textbf{Models} We experiment with finetuning \texttt{Qwen2.5-7B} and \texttt{Qwen2.5-7B-Instruct}. To produce AI judgments  or ground truth responses, we use \texttt{Qwen2.5-72B-Instruct} unless stated otherwise.

\textbf{Training} We finetune the model for 2 epochs using DPO with a batch size of 1024 and a maximum sequence length of 2048. We use a cosine learning rate schedule with a max LR of 3e-6 and a min LR of 2e-6.\footnote{When training models with Ultrafeedback, we instead used a minimum learning rate of 3e-7. We found this parameter resulted in a slightly stronger baseline when learning from this feedback.}  We use OpenRLHF for training \citep{Hu2024OpenRLHFAE}, and we train on one 8xH100 node with 80GB GPU memory, which took roughly 3 hours for each model.

\textbf{Benchmark Data} We evaluate our method on five benchmarks: IFEval \citep{Zhou2023InstructionFollowingEF}, InFoBench \citep{qin2024infobench}, FollowBench \citep{Jiang2023FollowBenchAM}, AlpacaEval \citep{AlpacaEval}, and Arena-Hard \citep{ArenaHard}. The first three of these measure instruction following ability in the presence of fine-grained constraints. The last two measure ``general-purpose'' instruction following ability, using naturalistic instructions based on user queries collected in the wild.

% \gncomment{Your method is not limited to WildChat, so it seems like either (1) this can be treated as an experimental detail and moved to the experiments section, or (2) you should emphasize more strongly that your work is applicable elsewhere as well. I'd prefer (1), but if it'd be too difficult you can do (2) as well.} To begin, we need a source of instructions. Our method is compatible with any  we use WildChat, a source of natural conversations between users and AI language models crowdsourced from users across the world \citep{Zhao2024WildChat1C}. We filter out conversations with toxic content, conversations tagged as being in a language other than English, and conversations consisting of greater than two conversational turns.

\subsection{Baselines}
%\gncomment{I think you should move this into the following section and re-name it ``Experimental Setup and Results''.}
To show that RLCF is more effective than existing approaches, we compare against baselines: \textit{instruction finetuning}, \textit{specially-trained reward models} (using either a single reward or mixture of rewards), and \textit{prompted AI judges} (using either a single evaluation rubric or a mixture of rubrics).

\noindent \textbf{Instruction Finetuning:} We compare with instruction finetuning, to isolate the benefit of additional knowledge from the manner it is given (ground truth or rewards). Here, we distill \citep{Hinton2015DistillingTK} from a larger model, \texttt{Qwen2.5-72B-Instruct}, finetuned via \texttt{LlamaFactory} \citep{llamafactory}.

\noindent \textbf{Reward Models:}
We mirror our training approach for learning from checklist feedback, but using state-of-the-art reward models to decide which response should be chosen or rejected. We use \texttt{Skywork/Skywork-Reward-Gemma-2-27B} \citep{Liu2024SkyworkRewardBO} and \texttt{ArmoRM-Llama3-8B-v0.1} \citep{Wang2024InterpretablePV} -- both are highly rated on RewardBench \citep{Lambert2024RewardBenchER}.\footnote{  \texttt{Skywork/Skywork-Reward-Gemma-2-27B} and \texttt{ArmoRM-Llama3-8B-v0.1} are ranked as \#4 and \#24, respectively, on RewardBench as of July 2025.}

\noindent \textbf{Prompted AI Judge:}
We compare against using the same ``teacher'' model as a judge, without using rubrics. 
We query this teacher in two settings: 1) \textit{``Ultrafeedback''}, where the judge rates all candidate responses from 1-5 \citep{Cui2023UltraFeedbackBL} separately across four quality aspects (instruction following, helpfulness, truthfulness, honesty) and averages these scores; and 2) \textit{AI Judge}, where a near-identical prompt as RLCF is used (\S\ref{sec:scoring_with_judge}) to similarly sample 25 scores between 0 and 100 from the judge.
% where the LM judge is given all candidate responses and simultaneously judges them from 1 to 5 \citep{Cui2023UltraFeedbackBL}. This procedure is performed separately for four aspects of response quality (instruction following, helpfulness truthfulness, and honestly), and then these scores are averaged.
%Second, we closely follow the same prompt that we use when scoring checklists via AI judge described in \S\ref{sec:scoring_with_judge} by providing a concrete rubric, requesting a score between 0 and 100, and taking the average of 25 sampled scores from the judge. 

\begin{figure}
    
\centering
\includegraphics[trim={2cm 8.2cm 31cm 20cm}, clip, width=0.78\linewidth]{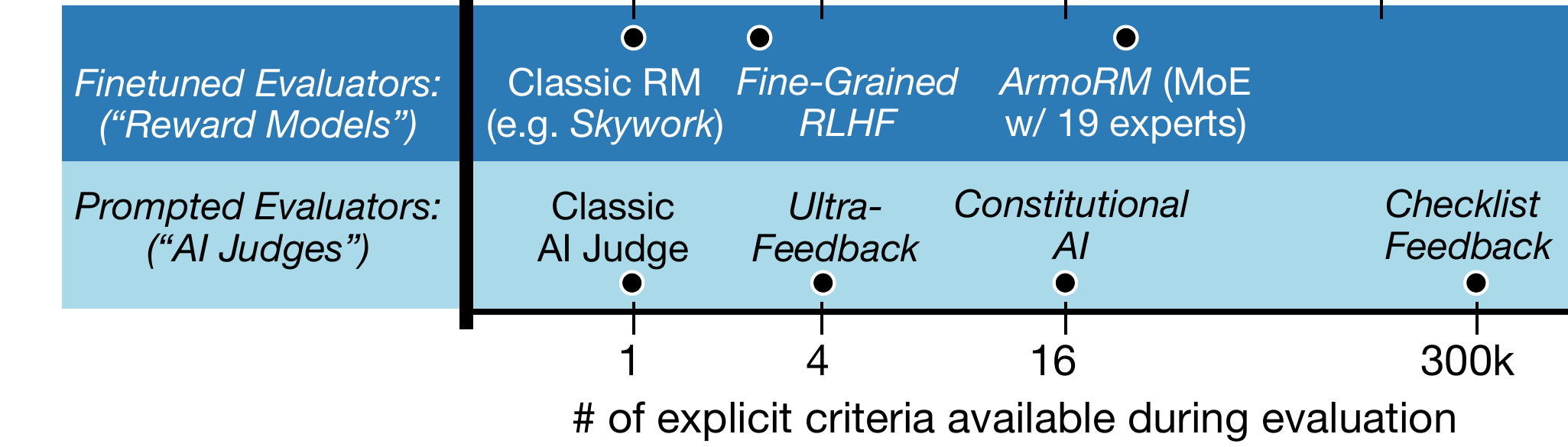}
  \vspace{-5pt}
  \caption{Checklist feedback can be viewed as an extreme mixture-of-evaluators, where the space of (prompted) evaluators is unbounded and a unique subset of evaluators is chosen for each instruction.}
  \label{fig:taxonomy_of_evaluators}
\vspace{8pt}
\end{figure}

In \autoref{fig:taxonomy_of_evaluators}, we unify these methods of automatic evaluation to distinguish our method from prior art. In this context, checklist feedback can be viewed as a very large mixture of prompted evaluators.

\section{Results}
\label{results}

\subsection{RL from Checklist Feedback consistently improves language models}

\begin{table}[t]
\centering
\begin{tabular}{l  c c  c  c | c | c c c }
\toprule
& \multicolumn{2}{c}{IFEval (prompt)} &  \multicolumn{2}{c}{IFEval (inst.)} &  &  \multicolumn{3}{c}{InFoBench}
\\
 &  \textbf{Loose} &  \textbf{Strict} &  \textbf{Loose} &  
\textbf{Strict} & \textbf{Avg} & \textbf{Easy} & \textbf{Hard}  & \textbf{Overall} \\
\midrule
GPT-4 & \textcolor{neutral}{79.3} & \textcolor{neutral}{76.9} & \textcolor{neutral}{85.4} & \textcolor{neutral}{83.6} & \textcolor{neutral}{81.3} & \textcolor{neutral}{89.3} & \textcolor{neutral}{86.4} & \textcolor{neutral}{87.3} \\
\midrule
\textit{Qwen2.5-7B-Instruct} &  \textcolor{neutral}{75.0} & \textcolor{neutral}{72.5} & \textcolor{neutral}{81.8} & \textcolor{neutral}{79.9} &  \textcolor{neutral}{77.3}  & \textcolor{neutral}{82.7} & \textcolor{neutral}{76.0} & \textcolor{neutral}{78.1}  \\
+ SFT (Distilled) & \textcolor{negative}{66.9} & \textcolor{negative}{64.1} & \textcolor{negative}{75.3} & \textcolor{negative}{72.8} & \textcolor{negative}{69.8} & \textcolor{negative}{79.9} & \textcolor{negative}{70.6} & \textcolor{negative}{73.5}\\
+ DPO (via Skywork) & \textcolor{positive}{75.8} & \textcolor{negative}{68.0} & \textcolor{positive}{83.2} & \textcolor{negative}{78.5} & \textcolor{negative}{76.0} & \textcolor{negative}{81.0} & \textcolor{positive}{82.4} & \textcolor{positive}{82.0} \\
+ DPO (via ArmoRM) & \textcolor{negative}{73.8} & \textcolor{negative}{70.2} & \textcolor{neutral}{81.7} & \textcolor{negative}{78.3} & \textcolor{negative}{76.0} & \textcolor{positive}{\textbf{84.2}} & \textcolor{positive}{83.1} & \textcolor{positive}{83.5} \\
+ DPO (via Ultrafbk.) & \textcolor{negative}{71.5} & \textcolor{negative}{69.1} & \textcolor{negative}{79.9} & \textcolor{negative}{77.7} & \textcolor{negative}{74.6} & \textcolor{neutral}{82.3} & \textcolor{positive}{79.0} & \textcolor{positive}{80.0} \\
+ DPO (via AI Judge) & \textcolor{negative}{73.0} & \textcolor{negative}{68.9} & \textcolor{negative}{80.9} & \textcolor{negative}{77.8} & \textcolor{negative}{75.2} & \textcolor{negative}{81.0} & \textcolor{negative}{73.9} & \textcolor{negative}{76.1} \\
+ DPO (RLCF) & \textcolor{positive}{\textbf{77.3}} & \textcolor{neutral}{72.6} & \textcolor{positive}{\textbf{84.1}} & \textcolor{neutral}{80.3} & \textcolor{positive}{\textbf{78.6}} & \textcolor{positive}{\textbf{84.2}} & \textcolor{positive}{\textbf{84.0}} & \textcolor{positive}{\textbf{84.1}} \\
\midrule
\textit{Qwen2.5-7B (Base)} & \textcolor{neutral}{35.7} & \textcolor{neutral}{30.5} & \textcolor{neutral}{46.6} & \textcolor{neutral}{42.1} & \textcolor{neutral}{38.7} & \textcolor{neutral}{68.8} & \textcolor{neutral}{77.4} & \textcolor{neutral}{74.8} \\
+ SFT on WildChat & \textcolor{positive}{38.1} & \textcolor{positive}{33.5} & \textcolor{positive}{52.2} & \textcolor{positive}{48.6} & \textcolor{positive}{43.1} & \textcolor{positive}{78.1} & \textcolor{positive}{80.1} & \textcolor{positive}{79.5}\\
+ DPO (RLCF) & \textcolor{positive}{\textbf{43.4}} & \textcolor{positive}{\textbf{35.9}} & \textcolor{positive}{\textbf{56.4}} & \textcolor{positive}{\textbf{49.2}} & \textcolor{positive}{46.2} & \textcolor{positive}{\textbf{80.6}} & \textcolor{positive}{\textbf{80.5}} & \textcolor{positive}{\textbf{80.5}} \\
\midrule
\end{tabular}
\vspace{3pt}
\caption{For instruction following, RLCF leads to large gains with open-ended constraints (InFoBench) and slightly positive or neutral changes on format-based constraints (IFEval). Off-the-shelf reward models help on InFoBench but hurt on IFEval. We show positive results (relative to the baseline) in \textcolor{positive}{blue}, negative in \textcolor{negative}{orange}, and neutral (within 0.5) in \textcolor{neutral}{gray}; the top variant of a given model is bolded.}
\label{tab:ifeval}
\end{table}

\begin{table}[t]
\centering
\fontsize{7.3}{8.3}\selectfont

\begin{tabular}{l | c c c c c c | c c c c c c | c}
\toprule
\textit{FollowBench} & \multicolumn{6}{|c}{\textit{Soft Satisfaction Rate}}  & \multicolumn{6}{|c|}{\textit{Hard Satisfaction Rate}} &  \\
\textbf{Level} &  \textbf{L1} &   \textbf{L2} &   \textbf{L3} &   \textbf{L4} &   \textbf{L5} &  \textbf{Avg} &   \textbf{L1} &   \textbf{L2} &   \textbf{L3} &   \textbf{L4} &   \textbf{L5} &  \textbf{Avg} &   \textbf{CSL}  \\
\midrule
GPT-4 &   \textcolor{neutral}{89.2}  & \textcolor{neutral}{89.3} & \textcolor{neutral}{87.6} & \textcolor{neutral}{88.1} & \textcolor{neutral}{84.9} & \textcolor{neutral}{87.8} & \textcolor{neutral}{89.2} & \textcolor{neutral}{87.6} & \textcolor{neutral}{83.6} & \textcolor{neutral}{83.0} & \textcolor{neutral}{75.1} & \textcolor{neutral}{83.7} & \textcolor{neutral}{3.52} \\
\midrule
\textit{Qwen2.5-7B-Instr.} &   \textcolor{neutral}{87.4}  & \textcolor{neutral}{84.0} & \textcolor{neutral}{83.0} & \textcolor{neutral}{79.6} & \textcolor{neutral}{79.0} & \textcolor{neutral}{82.6} & \textcolor{neutral}{87.4} & \textcolor{neutral}{80.6} & \textcolor{neutral}{72.3} & \textcolor{neutral}{62.2} & \textcolor{neutral}{54.4} & \textcolor{neutral}{71.4} & \textcolor{neutral}{3.05}
  \\
+ SFT (Distilled) & \textcolor{neutral}{87.5} & \textcolor{negative}{83.2} & \textcolor{positive}{84.4} & \textcolor{negative}{76.8} & \textcolor{negative}{74.9} & \textcolor{negative}{81.4} & \textcolor{neutral}{87.5} & \textcolor{negative}{78.3} & \textcolor{positive}{73.9} & \textcolor{negative}{60.7} & \textcolor{negative}{49.1} & \textcolor{negative}{69.9} & \textcolor{negative}{2.90} \\
+ DPO (Skywork) & \textcolor{negative}{79.6} & \textcolor{negative}{84.1} & \textcolor{negative}{77.7} & \textcolor{negative}{77.7} & \textcolor{negative}{78.1} & \textcolor{negative}{79.4} & \textcolor{negative}{79.6} & \textcolor{neutral}{81.1} & \textcolor{negative}{67.4} & \textcolor{positive}{62.9} & \textcolor{negative}{56.5} & \textcolor{negative}{69.5} & \textcolor{negative}{2.88} \\
+ DPO (ArmoRM) & \textcolor{negative}{86.4} & \textcolor{positive}{84.6} & \textcolor{negative}{79.1} & \textcolor{negative}{79.2} & \textcolor{negative}{76.9} & \textcolor{negative}{81.2} & \textcolor{negative}{86.4} & \textcolor{positive}{82.9} & \textcolor{negative}{69.0} & \textcolor{positive}{63.9} & \textcolor{positive}{49.7} & \textcolor{neutral}{70.4} & \textcolor{neutral}{3.10} \\
+ DPO (Ultrafbk.) & \textcolor{positive}{88.5} & \textcolor{neutral}{84.1} & \textcolor{neutral}{82.5} & \textcolor{negative}{76.3} & \textcolor{negative}{72.6} & \textcolor{negative}{80.8} & \textcolor{positive}{88.5} & \textcolor{neutral}{81.1} & \textcolor{negative}{62.4} & \textcolor{positive}{63.5} & \textcolor{neutral}{54.9} & \textcolor{positive}{72.6} & \textcolor{negative}{2.98} \\
+ DPO (AI Judge) & \textcolor{neutral}{87.2} & \textcolor{positive}{87.9} & \textcolor{negative}{75.7} & \textcolor{neutral}{79.2} & \textcolor{negative}{77.6} & \textcolor{negative}{81.5} & \textcolor{neutral}{87.2} & \textcolor{positive}{83.5} & \textcolor{negative}{62.4} & \textcolor{positive}{63.5} & \textcolor{neutral}{54.9}& \textcolor{negative}{70.3}  & \textcolor{negative}{2.95} \\
DPO (RLCF) & \textcolor{positive}{\textbf{88.6}} & \textcolor{positive}{\textbf{88.8}} & \textcolor{positive}{\textbf{83.8}} & \textcolor{positive}{\textbf{79.9}} & \textcolor{positive}{\textbf{81.0}} & \textcolor{positive}{\textbf{84.4}} & \textcolor{positive}{\textbf{88.6}} & \textcolor{positive}{\textbf{85.2}} & \textcolor{positive}{\textbf{75.8}} & \textcolor{positive}{\textbf{65.1}} & \textcolor{positive}{\textbf{61.8}} & \textcolor{positive}{\textbf{75.3}} & \textcolor{positive}{\textbf{3.30}} \\
\midrule
\textit{Qwen2.5-7B (Base)} & \textcolor{neutral}{55.9} & \textcolor{neutral}{60.7} & \textcolor{neutral}{56.6} & \textcolor{neutral}{56.1} & \textcolor{neutral}{54.6} & \textcolor{neutral}{56.8} & \textcolor{neutral}{55.9} & \textcolor{neutral}{49.1} & \textcolor{neutral}{36.1} & \textcolor{neutral}{33.4} & \textcolor{neutral}{19.5} & \textcolor{neutral}{38.8} & \textcolor{neutral}{1.20} \\
+ SFT (WildChat) & \textcolor{positive}{65.4} & \textcolor{positive}{75.3} & \textcolor{positive}{71.6} & \textcolor{positive}{64.7} & \textcolor{positive}{\textbf{65.1}} & \textcolor{positive}{68.4} & \textcolor{positive}{65.4} & \textcolor{positive}{69.2} & \textcolor{positive}{\textbf{57.4}} & \textcolor{positive}{\textbf{46.9}}& \textcolor{positive}{40.3} & \textcolor{positive}{55.8} & \textcolor{positive}{2.02} \\
+ DPO (RLCF) & \textcolor{positive}{70.6} & \textcolor{positive}{76.0} & \textcolor{positive}{69.5} & \textcolor{positive}{63.6} & \textcolor{positive}{57.8} & \textcolor{positive}{67.5} & \textcolor{positive}{70.6} & \textcolor{positive}{67.7} & \textcolor{positive}{49.6} & \textcolor{positive}{42.4} & \textcolor{positive}{28.3} & \textcolor{positive}{51.7} & \textcolor{positive}{2.08} \\
+ RLCF w/o code & \textcolor{positive}{\textbf{70.9}} & \textcolor{positive}{\textbf{77.1}} & \textcolor{positive}{\textbf{73.3}} & \textcolor{positive}{\textbf{66.0}} & \textcolor{positive}{63.5} & \textcolor{positive}{\textbf{70.2}} & \textcolor{positive}{\textbf{70.9}} &\textcolor{positive}{ \textbf{70.0}} & \textcolor{positive}{56.5} & \textcolor{positive}{42.9} & \textcolor{positive}{\textbf{36.3}} & \textcolor{positive}{\textbf{55.3}} & \textcolor{positive}{\textbf{2.20}} \\
\bottomrule
\end{tabular}
\vspace{3pt}
\caption{RLCF leads to improvements on FollowBench  \textbf{across all metrics} when starting with an instruction-tuned model, while using an off-the-shelf reward model for preference labeling leads to regressions for most metrics. This algorithm also helps when applied to a non-instruction-tuned model, though it does not beat supervised finetuning. ``CSL'' stands for ``Constraint Satisfaction Level''. We show positive results (relative to the baseline) in \textcolor{positive}{blue}, negative in \textcolor{negative}{orange}, and neutral (within 0.5 for satisfaction rate or 0.05 for CSL) in \textcolor{neutral}{gray}; the top variant of a given model is bolded.}
\label{tab:followbench}
\end{table}

Our proposed approach, RLCF, demonstrates consistent gains across all benchmarks (\autoref{tab:ifeval}, \autoref{tab:followbench}, and \autoref{tab:alpaca}). On IFEval's ``loose'' metrics (which apply minor preprocessing to responses before checking for correctness), RLCF improves \texttt{Qwen-7B-Instruct} by 2.8-3.0\% (relative), as shown in the left half of \autoref{tab:ifeval}. On FollowBench (shown in \autoref{tab:followbench}), RLCF achieves an 8.2\% increase on Constraint Satisfaction Level (CSL; the expected proportion of constraints satisfied) and a 5.5\% increase on average Hard Satisfaction Rate (how often all constraints are satisfied). RLCF also performs competitively on InFoBench (right half of \autoref{tab:ifeval}), achieving results comparable to the best-performing reward model-based approaches while maintaining consistent gains across all constrain-based benchmarks.
%We also see sizeable gains on Infobench (right half of \autoref{tab:ifeval}, slightly outperforming reinforcement learning from reward model-based rewards (which is also effective on this benchmark). 
On ``general use-case'' instruction following benchmarks, 
%which are widely used for language model evaluation, 
RLCF consistently increases the win rate of \texttt{Qwen2.5-7B} over \texttt{GPT-4} (shown in \autoref{tab:alpaca}), with the relative improvement ranging from 2.8\% to 8.4\%.

\begin{table}[t]
\centering
\small
\begin{tabular}{l  c c  c c }
\toprule
&  \multicolumn{2}{c}{\textit{Arena-Hard}} &   \multicolumn{2}{c}{\textit{AlpacaEval}}  \\
\midrule
&  \textbf{Vanilla} &  \textbf{Style-Controlled} &   \textbf{Vanilla} &  \textbf{Length-Controlled}  \\
\midrule
GPT-4 (0314) &    \textcolor{baseline}{50.0} & \textcolor{baseline}{50.0} & \textcolor{baseline}{22.1} & \textcolor{baseline}{35.3} \\
\midrule
\textit{Qwen2.5-7B-Instruct} &    \textcolor{baseline}{51.3} & \textcolor{baseline}{42.8} & \textcolor{baseline}{33.5} & \textcolor{baseline}{36.2} \\
+ SFT (Distilled) & \textcolor{negative}{32.6} & \textcolor{negative}{29.2} & \textcolor{positive}{36.1} & \textcolor{negative}{33.3} \\
+ DPO (via Skywork)& \textcolor{positive}{\textbf{55.1}} & \textcolor{positive}{\textbf{50.3}} & \textcolor{positive}{\textbf{44.8}} & \textcolor{positive}{\textbf{41.5}} \\
+ DPO (via ArmoRM) & \textcolor{neutral}{50.8} & \textcolor{positive}{46.4} & \textcolor{positive}{37.6} & \textcolor{positive}{38.1}\\
+ DPO (via Ultrafeedback) & \textcolor{positive}{52.8} & \textcolor{positive}{47.9} & \textcolor{neutral}{33.7} & \textcolor{positive}{38.7} \\
+ DPO (via AI Judge) & \textcolor{neutral}{51.0} & \textcolor{positive}{44.4} & \textcolor{negative}{28.8} & \textcolor{negative}{33.4} \\
+ DPO (RLCF) &  \textcolor{positive}{54.6} & \textcolor{positive}{48.4} & \textcolor{positive}{36.2} & \textcolor{positive}{37.1} \\
\midrule
\textit{Qwen2.5-7B (Base)} & \textcolor{baseline}{19.6} & \textcolor{baseline}{24.1} & \textcolor{baseline}{8.9} & \textcolor{baseline}{9.4} \\
+ SFT on WildChat & \textcolor{negative}{8.8} & \textcolor{negative}{8.8} & \textcolor{neutral}{9.4} & \textcolor{negative}{7.5} \\
+ DPO (RLCF) & \textcolor{neutral}{19.4}  & \textcolor{negative}{21.6} & \textcolor{positive}{\textbf{11.2}} & \textcolor{positive}{10.5}  \\
+ RLCF w/o program verification & \textcolor{positive}{\textbf{23.1}} & \textcolor{positive}{\textbf{27.1}} & \textcolor{positive}{11.0} & \textcolor{positive}{\textbf{13.9}}  \\
\bottomrule
\end{tabular}
\vspace{3pt}
\caption{We compare methods on two ``general'' instruction following benchmarks: Arena-Hard and AlpacaEval. RLCF gives modest but consistent gains on both the original metric and length/style-controlled metric on each benchmark. We show positive results (relative to the baseline) in \textcolor{positive}{blue}, negative in \textcolor{negative}{orange}, and neutral (within 0.5) in \textcolor{neutral}{gray}; the top variant of a given model is bolded.}
\label{tab:alpaca}
\vspace{-10pt}
\end{table}

\subsection{Comparing automatic evaluators}
\label{sec:comparing_evaluators}

In \autoref{tab:ifeval}, \autoref{tab:followbench}, and \autoref{tab:alpaca}, we observe our approach of performing RL from Checklist Feedback (RLCF) outperforms RL from other sources of automatic evaluation across most benchmarks. However, off-the-shelf reward models show mixed results depending on the benchmark. \textit{Skywork} (\texttt{Skywork-Reward-Gemma-2-27B}), a leading model on the RewardBench leaderboard, shows strong improvements with RLHF on InFoBench, Arena-Hard, and AlpacaEval -- RLHF via Skywork notably outperforms RLCF on AlpacaEval by a large margin. However, Skywork-guided RLHF leads to notable regressions as on IFEval and FollowBench. Similarly, RLHF with \textit{ArmoRM} shows significant improvements on AlpacaEval and InFoBench, modest/mixed results on Arena-Hard and FollowBench, and significant regressions on IFEval.

\begin{table}[h]
\centering

% \gncomment{Given that this is largely a negative result, do we want to feature it in the limited space we have in the paper? I think it's still interesting, but if we need to move something to an appendix, we could move this.}
%\renewcommand{\arraystretch}{0.95}

\begin{tabular}{r | c c c c c}
\toprule
\small
& Chat & Chat Hard & Safety & Reasoning \\
\midrule
Skywork-27B & 96.1 & \textbf{89.9} & \textbf{93.0} & \textbf{98.1} \\
ArmoRM & \textbf{96.9} & 76.8 & 90.5 & 97.3 \\
Checklist-Based Reward & 90.0 & 80.7 & 71.4 & 88.5 \\
\bottomrule
\end{tabular}
\vspace{3pt}
\caption{On RewardBench, Specialized reward models like Skywork-27B and ArmoRM excel at predicting which response is superior. Our checklist-based approach is worse on this this benchmark, but still achieves competitive performance on challenging categories like Chat Hard and Reasoning.}
\label{tab:rewardbench}
\vspace{-10pt}
\end{table}

We also evaluate checklist feedback's ability as a judge on RewardBench\footnote{Unlike our method for checklist generation on WildChat, here we do not use any ground truth or output from other models when generating checklists.}. \autoref{tab:rewardbench} shows checklist scores are well-correlated with preference annotations on RewardBench, especially for the "Chat" and "Chat Hard" categories \citep{Lambert2024RewardBenchER}. However, specialized reward models (\textit{Skywork}, \textit{ArmoRM}), are much better here, though worse at supervising RLHF. This follows prior evidence that reward model ``accuracy'' is poorly correlated with efficacy in RLHF \citep{Malik2025RewardBench2A, Razin2025WhatMA}.
Note that our method shows relatively less correlation with Safety -- our implementation of RLCF is not designed as a substitute for safety alignment (see more discussion in \S\autoref{sec:other_tasks}).

\subsection{Learning from candidate-based vs directly-generated checklists}
\label{sec:comparing_req_gen_methods}
In \autoref{sec:checklist_generation}, we described a novel method for \textit{candidate-based} checklist generation, and we presented some intrinsic evaluation showing that this method generates good checklists. Do these checklists indeed translate to better models after RL training?

\begin{table}[t]
\centering
\small
\begin{tabular}{l  c c  c  c | c | c | c c }
\toprule
& \multicolumn{2}{c}{IFEval (prompt)} &  \multicolumn{2}{c}{IFEval (inst.)} &  &  InFoBench  & \multicolumn{2}{c}{FollowBench} 
\\
 &  \textbf{Loose} &  \textbf{Strict} &  \textbf{Loose} &  
\textbf{Strict} & \textbf{Avg} & \textbf{Overall} & \textbf{SSR} & \textbf{ HSR} \\
\midrule
\textit{Qwen2.5-7B-Instruct} &  \textcolor{neutral}{75.0} & \textcolor{neutral}{72.5} & \textcolor{neutral}{81.8} & \textcolor{neutral}{79.9} &  \textcolor{neutral}{77.3}  & \textcolor{neutral}{78.1}  & \textcolor{neutral}{82.6} &  \textcolor{neutral}{71.4}\\
+ RLCF (direct) & \textcolor{negative}{74.3} & \textcolor{negative}{69.5} & \textcolor{neutral}{81.5} & \textcolor{negative}{77.9} & \textcolor{negative}{76.9} & \textcolor{positive}{\textbf{84.3}} & \textcolor{neutral}{82.5} & \textcolor{positive}{72.8}  \\
+ RLCF (candidate-based) & \textcolor{positive}{\textbf{77.3}} & \textcolor{neutral}{\textbf{72.6}} & \textcolor{positive}{\textbf{84.1}} & \textcolor{positive}{\textbf{80.3}} & \textcolor{positive}{\textbf{78.6}} & \textcolor{positive}{84.1} & \textcolor{positive}{\textbf{84.4}} & \textcolor{positive}{\textbf{75.3}}  \\
\midrule
\end{tabular}
\vspace{3pt}
\caption{Using candidate-based checklists is crucial to making RLCF work, suggesting that the quality and properties of checklists are important for learning from checklist feedback.}
\label{tab:comparing_req_gen_methods}
\end{table}

In \autoref{tab:comparing_req_gen_methods}, we observe that RLCF is consistently better using ``candidate-based'' checklists than using checklists generated ``directly`` by prompting: 2\% better on IFEval, equally good on InFoBench, and 2-3\% better on FolllowBench. This shows that RLCF depends on detailed, and objective checklists that may offer more new information than checklists obtained directly from the original prompt.

\subsection{Where does checklist feedback help?}
\label{sec:bucket_analysis}

\begin{table}[h]
\centering

\begin{tabular}{l | c | c c c c}
\toprule
& Avg (HSR) & Format & Style & Situation & Content \\
\midrule
GPT-4 & \textcolor{neutral}{83.7} & \textcolor{neutral}{83.3} & \textcolor{neutral}{97.3} & \textcolor{neutral}{78.2} & \textcolor{neutral}{76.0} \\
\midrule
\textit{Qwen2.5-7B-Instruct} & \textcolor{neutral}{71.4} & \textcolor{neutral}{60.0} & \textcolor{neutral}{87.3} & \textcolor{neutral}{78.1} & \textcolor{neutral}{60.0} \\

+ DPO (Skywork) & \textcolor{negative}{69.5} & \textcolor{positive}{62.7} & \textcolor{positive}{88.0} & \textcolor{negative}{74.7} & \textcolor{negative}{52.8} \\
+ DPO (ArmoRM) & \textcolor{negative}{70.4} & \textcolor{positive}{62.0} & \textcolor{positive}{89.3} & \textcolor{negative}{71.8} & \textcolor{negative}{58.4} \\
+ SFT (Distilled) & \textcolor{neutral}{71.1} & \textcolor{positive}{61.3} & \textcolor{negative}{85.3} & \textcolor{positive}{80.0} & \textcolor{negative}{57.6} \\
+ RLCF \textit{w/o prompt-based scoring} & \textcolor{negative}{73.6} & \textcolor{positive}{62.7} & \textcolor{positive}{90.7} & \textcolor{positive}{\textbf{81.8}} & \textcolor{negative}{59.2} \\
+ RLCF \textit{w/o program verification}) & \textcolor{positive}{73.8} & \textcolor{positive}{\textbf{68.7}} & \textcolor{positive}{\textbf{91.3}} & \textcolor{positive}{80.0} & \textcolor{negative}{55.2} \\
+ RLCF & \textcolor{positive}{\textbf{75.3}} & \textcolor{positive}{64.0} & \textcolor{positive}{90.7} & \textcolor{positive}{80.0} & \textcolor{positive}{\textbf{66.4}} \\
\bottomrule
\end{tabular}
\vspace{3pt}
  \caption{On FollowBench, RLCF helps especially with ``content'' constraints, which are qualifiers that restrict the valid space of answers. The metric shown is ``average hard satisfaction rate''. We speculate that RLCF helps models attend to full instructions. We show positive results in \textcolor{positive}{blue}, negative in \textcolor{negative}{orange}, and neutral (within 0.5) in \textcolor{neutral}{gray}; the top variant of a given model is bolded.
  %\gncomment{Usually I expect the proposed method to be at the bottom, maybe we could put the baselines on top, perhaps in a separate section?}
  }
\label{tab:followbench_bucket_analysis}
\vspace{-10pt}
\end{table}

Does checklist feedback help with a specific aspect of instructions, such as rule-based format constraints? Performance on specific constraint types on FollowBench, shown in \autoref{tab:followbench_bucket_analysis}, shows that, unsurprisingly, prompt-based scoring is helpful for prompts involving style or format constraints. We also see that \textbf{RLCF is best for ``content'' constraints}, which are qualifiers included on open-ended questions to limit the valid space of answers (e.g. ``\textit{How might economic data from \emph{the past quarter} affect the Fed's decision on interest rates? Additionally, consider \emph{how inflation rates might influence their decision.}}''). This suggests \textbf{checklist feedback incentivizes models to attend to the full instruction} rather than using a few influential spans to generate responses.

\begin{table}[t]
\centering
\footnotesize

\newcommand{\promptrow}[1]{\rowcolor{lightgray!40}
\textbf{Prompt} & \multicolumn{5}{l}{#1} \\
\arrayrulecolor{white}\midrule}

\newcommand{\checklistrow}[1]{\rowcolor{lightgray!40}
\textbf{Checklist} & \multicolumn{5}{l}{\parbox{0.85\textwidth}{#1}} \\
\arrayrulecolor{black}\midrule}

\newcommand{\titlerow}{
\multicolumn{2}{l|}{\textbf{Responses}} & \textbf{Skywork Reward} & \textbf{AI Judge} & \textbf{Checklist (code)} & \textbf{Checklist (judge)} \\
}

\newcommand{\responsecell}[1]{
\multicolumn{2}{l|}{{\parbox{0.5\textwidth}{``#1''}}} }

\newcommand{\weightmacro}[1]{
\texttt{\scriptsize(weight: #1/100)}}

\begin{tabularx}{\textwidth}{
    p{0.08\textwidth}  
    p{0.43\textwidth} |% Responses
    p{0.08\textwidth}   |% Skywork
    p{0.08\textwidth}   |% AI Judge
    p{0.09\textwidth}   |% Checklist (judge)
    p{0.09\textwidth}   % Checklist (code)
}
\toprule
\titlerow

\midrule\midrule
\promptrow{Translate to Spanish: ``Hello how are you doing?''} 
\checklistrow{
1. Is the generated text in Spanish? 
\weightmacro{100}\\
2. Is the text an accurate and complete translation of the English sentence? \weightmacro{100}
}

\responsecell{¡Hola, ¿cómo estás?}
& 25.5 & 100.0 & 100.0 & 95.2 \\
\midrule

\responsecell{HOLA, ¿CÓMO TE ESTÁScaller"H!impo-rtant"Endpoint unfinished\begin{CJK}{UTF8}{gbsn}际\end{CJK}"$>$vak dao ''`\begin{CJK}{UTF8}{gbsn}圣诞`"[...]\end{CJK}}
& 0.0 & 100.0 & 100.0 & 0.0 \\

\midrule
\midrule
%%%%%%%%%%%%%%%%%%%%%%%%%%%%%%
 \promptrow{make a sentence with ``dense''} 
 \checklistrow{
 1. Does the generated text contain the word ``dense''? \weightmacro{100} \\
 2. Is the generated text a coherent and grammatically correct sentence? \weightmacro{75}}

 \responsecell{The forest was dense, with trees so close together [...]}
 & 33.1 & 100.0  & 100.0 & 97.3 \\
 \midrule

 \responsecell{The forest was blanketed by a layer of dense vegetation.}
 & 8.0 & 100.0 & 100.0 & 96.6 \\

 % \midrule
 % \midrule
%%%%%%%%%%%%%%%%%%%%%%%%%%%%%%%%%%%%

% \promptrow{How does dyspraxia affect the delivery of incident command in the fire service} 
% \checklistrow{
% 1. Does the text explain how dyspraxia affects [...]\weightmacro{100} \\
% 2. Does the text have a logical flow of information? \weightmacro{90} \\
% 3. Does the text cover multiple ways dyspraxia impacts incident command?\weightmacro{80} \\
% 4. Is the text free of irrelevant information?\weightmacro{75} \\
% 5. Is the text concise \weightmacro{75}}

% \responsecell{Dyspraxia, also known as developmental coordination disorder (DCD), can significantly impact the delivery of incident command [...]}
% & 77.2 & 100.0 & N/A & 83.6 \\
% \midrule

% \responsecell{Dyspraxia, also known as developmental\begin{CJK}{UTF8}{gbsn}协调与改进：您提到的是韵律、发音和句子\end{CJK} [...]}
% & 0.0 & 0.0 & N/A & 13.6 \\
\bottomrule

\end{tabularx}
\vspace{3pt}
\caption{Comparing the scores assigned to various prompts and responses, we see that reward models are too sensitive, prompted AI judges are too granular, and checklists give stable, interpretable scores.}
\label{tab:model-comparison-enhanced}
\end{table}

This hypothesis is supported by qualitative analysis of the preference data in \autoref{tab:model-comparison-enhanced}.
We observe that using an AI judge with a single rubric is often insensitive to major changes in the prompt. When the user asks to translate an utterance to Spanish, the AI judge assigns a 100-point score to both a perfect response and a poor response that contains a slightly flawed translation along with incoherent phrases. In the second example, Skywork-27B assigns wildly different scores to responses with identical meaning.
The two scoring components of checklist feedback -- a verification program and a checklist-based AI judge -- can balance each other's shortcomings, as shown in the first example.

\subsection{Does RLCF lead to specialization at the expense of generality?}
\label{sec:other_tasks}

\begin{table}[t]
\centering
\begin{tabular}{l  c c  c  | c | c c }
\toprule
& \multicolumn{3}{c|}{XSTest} & GSM8K & \multicolumn{2}{c}{TruthfulQA} 
\\
 & Safe ($\uparrow$) & Unsafe  ($\uparrow$) & Overall  ($\uparrow$) & Accuracy & MC1  & MC2 \\
\midrule
\textit{Qwen2.5-7B-Instruct} &  \textcolor{neutral}{92.0} & \textcolor{neutral}{83.0} &  \textcolor{neutral}{86.0}  & \textcolor{neutral}{83.2}  & \textcolor{neutral}{43.5} &  \textcolor{neutral}{60.4}\\
+ RLCF & \textcolor{positive}{\textbf{95.6}} & \textcolor{negative}{81.0} & \textcolor{positive}{86.9} & \textcolor{negative}{82.2} & \textcolor{negative}{42.0} & \textcolor{negative}{59.0}  \\
+ DPO (Skywork) & \textcolor{negative}{\textbf{90.4}} & \textcolor{positive}{\textbf{89.0}} & \textcolor{positive}{\textbf{88.6}} & \textcolor{positive}{\textbf{85.6}} & \textcolor{positive}{\textbf{45.2}} & \textcolor{positive}{\textbf{63.1}}  \\
\midrule
\end{tabular}
\vspace{3pt}
\caption{In our experiments, RLCF uses prompts that focus primarily on daily assistance and writing \citep{Zhao2024WildChat1C}. This has a small effect on ``non-target'' tasks: refusal to answer unsafe prompts (measured on XSTest), basic mathematical reasoning (measured on GSM8K), and hallucination prevention (measured on TruthfulQA), suggesting that RLCF incurs modest trade-offs on underrepresented domains, which could be mitigated by improving domain coverage of training prompts.}
\label{tab:spec_vs_gen}
\vspace{-8pt}
\end{table}

Do the gains shown by RLCF on instruction following come at the expense of domains not well-represented in the training data (WildChat)? WildChat focuses primarily on daily assistance, advice, and analysis (75.5\%), with only 12\% on factual information and mathematics \citep{Zhao2024WildChat1C}.  Is this causing specialization at the expense of generality? We evaluate RLCF on three tasks with limited representation in WildChat: refusal to answer unsafe prompts (measured on XSTest \citep{xstest}), basic math (measured on GSM8K \citep{gsm8k}), and hallucination prevention (measured on TruthfulQA \citep{truthfulqa}). In \autoref{tab:spec_vs_gen}, we see that RLCF slightly alters the model's safety profile (reducing false refusals while slightly impairing true refusals) and reduces GSM8K and TruthfulQA performance by 1-1.5\%. This suggests a need for expanding \textit{WildChecklists} to a more diverse prompt distribution. Fortunately, this is easier than retraining a reward model. RMs like Skywork-27B show better generality out-of-the-box (e.g. in \autoref{tab:spec_vs_gen}) because they do not assume fixed criteria, but support for very different prompts would require retraining \citep{Malik2025RewardBench2A}.

\subsection{RLCF improves other model families off-policy}
We use \texttt{Qwen2.5} models as the policy and judge in our experiments. Does RLCF still work for other model families? Do our checklists capture universal criteria or are they tied to a particular choice of model? Using \textit{WildChecklists} with samples from \texttt{Qwen2.5-7B-Instruct} scored by \texttt{Qwen2.5-72B-Instruct}, we trained \texttt{Llama 3.1 8B Instruct} \citep{llama3} and \texttt{OLMo 2 7B Instruct} \citep{olmo2} off-policy with DPO. In \autoref{tab:off_policy_ifeval_infobench_followbench}, we see positive results with both models: \texttt{Llama} sees strong improvements with InFoBench/FollowBench after RLCF and \texttt{OLMo} sees strong improvements with IFEval after RLCF, and there are no regressions for either model. This suggests these checklists do capture universal criteria.

\begin{table}[t]
\centering
\small
\begin{tabular}{l  c c  c  c | c | c | c c }
\toprule
& \multicolumn{2}{c}{IFEval (prompt)} &  \multicolumn{2}{c}{IFEval (inst.)} &  &  InFoBench  & \multicolumn{2}{c}{FollowBench} 
\\
 &  \textbf{Loose} &  \textbf{Strict} &  \textbf{Loose} &  
\textbf{Strict} & \textbf{Avg} & \textbf{Overall} & \textbf{SSR} & \textbf{ HSR} \\
\midrule
\textit{Llama 3.1 8B Instruct} &  \textcolor{neutral}{79.1} & \textcolor{neutral}{71.3} & \textcolor{neutral}{83.5} & \textcolor{neutral}{76.2} &  \textcolor{neutral}{77.5}  & \textcolor{neutral}{83.1}  & \textcolor{neutral}{77.6} &  \textcolor{neutral}{68.0}\\
+ RLCF (off-policy) & \textcolor{positive}{\textbf{80.2}} & \textcolor{positive}{\textbf{71.9}} & \textcolor{neutral}{83.9} & \textcolor{neutral}{76.5} & \textcolor{positive}{\textbf{78.1}} & \textcolor{positive}{\textbf{84.2}} & \textcolor{positive}{\textbf{81.8}} & \textcolor{positive}{\textbf{72.3}}  \\
\midrule
\textit{OLMo 2 7B Instruct} &  \textcolor{neutral}{78.1} & \textcolor{neutral}{69.3} & \textcolor{neutral}{79.5} & \textcolor{neutral}{71.3} &  \textcolor{neutral}{74.6}  & \textcolor{neutral}{80.1}  & \textcolor{neutral}{72.3} &  \textcolor{neutral}{59.4}\\
+ RLCF (off-policy) & \textcolor{positive}{\textbf{79.6}} & \textcolor{positive}{\textbf{70.8}} & \textcolor{positive}{\textbf{82.0}} & \textcolor{positive}{\textbf{73.9}} & \textcolor{positive}{\textbf{76.6}} & \textcolor{neutral}{80.5} & \textcolor{neutral}{71.9} & \textcolor{positive}{\textbf{60.4}}  \\
\midrule
\end{tabular}
\vspace{3pt}
\caption{RLCF works \textit{off-policy} for  \texttt{OLMo 2 7B Instruct} and \texttt{Llama 3.1 8B Instruct}. For \texttt{Llama}, we see neutral results with IFEval and strong improvements on InFoBench and FollowBench. For \texttt{OLMo}, we see strong improvements with IFEval and neutral results on InFoBench and FollowBench.}
\label{tab:off_policy_ifeval_infobench_followbench}
\vspace{-11pt}
\end{table}

\subsection{How much compute is required for producing checklist-based AI judgments?}
\label{sec:num_judges}

\begin{wrapfigure}[13]{r}{0.44\textwidth}
  \centering
\vspace{2pt}
\includegraphics[trim={0.38cm 1.07cm 0.35cm 0.36cm}, clip, width=0.95\linewidth]{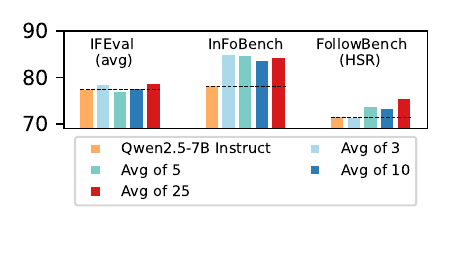}
  \caption{RLCF samples 25 scores when grading each requirement. This is expensive. Fortunately, much of the efficacy is retained using just 5 samples (55\% less clock time).}
  \label{fig:num_judges}
\end{wrapfigure}

As described in \autoref{sec:scoring_with_judge}, RLCF grades responses on each criterion using 25 samples from a judge model. This creates the computational bottleneck of our scoring procedure. In \autoref{fig:num_judges}, we evaluated models trained using RLCF modified to use fewer samples from the judge. Response grading on WildChat with 3, 5, 10, or 25 samples took 32, 40, 72, and 92 hours, respectively, on one 8xH100 node. We observe stable efficacy on IFEval\footnote{The models we trained all showed moderate variance on IFEval, so slight differences are likely due to noise.} and InFoBench across sample sizes. For FollowBench, using fewer samples hurts the ``content'' and ``situation'' categories, suggesting that a cheap, high-variance score may suffice for simpler criteria but not for difficult, ambiguous instructions.
%\include{qualitative_comparison}

% - ablate the effect of code vs no code
% - compare different design decisions using best-of-K reranking
% - consider providing RewardBench results, to show that RewardBench is perhaps not well-correlated with good training signals?

\section{Related Works}

%\gncomment{It'd be good to cover the more general methods of rule-conditioned reward or constitutional AI, which have global checklists \citep{glaese2022improving,Bai2022ConstitutionalAH}}
%\gncomment{You might want to expand on \citet{Cook2024TICKingAT} here a little more.}
% also need to cover Branch-Solve-Merge
%\gncomment{I just found this contemporaneous work, it's so recent that we probably don't need to treat it as prior work, but it might be interesting to look at \citep{chen2025rm}.}
Our method is a new means of generating synthetic AI feedback. This follows prior work that use ``AI feedback'' to guide reinforcement learning algorithms, either via a single prompt/rubric \citep{Tunstall2023ZephyrDD} or a collection of rubrics \citep{Cui2023UltraFeedbackBL}. In our paper, we show that checklist feedback is significantly more effective than UltraFeedback \citep{Cui2023UltraFeedbackBL}, which evaluates responses on four global principles. Our work is also related to prior works that use reward models as synthetic preference annotators for RL \citep{Sun2023SALMONSW}. In \autoref{tab:ifeval}, \autoref{tab:followbench}, \autoref{tab:alpaca}, we demonstrates the risks of using reward models to supervise RL \citep{Liu2024SkyworkRewardBO, ArmoRM}.

We focus on complex instruction following. One line of work synthesizes instructions with explicit, pathological constraints to train models to generalize to other instructions \citep{Xu2023WizardLMEL, He2024FromCT, conifer, autoif}. These papers use DPO on controlled candidate responses, while we contribute a drop-in evaluator which allows scoring responses sampled directly from a student model (opening the door to other RL algorithms like GRPO \citep{Shao2024DeepSeekMathPT}).

Our work is related to a nascent line of work that explores using rubrics for language model alignment and evaluation. \citet{Cook2024TICKingAT} demonstrate that using model-generated checklists can be useful at inference-time for proprietary LLMs. Similarly, \citet{Saha2023BranchSolveMergeIL} use generated checklists at inference time to improve constrained reasoning tasks. \citep{SaadFalcon2024LMUnitFE} use checklists to evaluate language models and match our finding that checklists can outperform reward models at response evaluation. \citet{Wang2025FromRT}, contemporaneously to us, introduce a rubric generation method called ``pre-comparison-derived criteria'' which also uses candidate responses sampled from different language models; they show that these criteria improve the agreement of automatic evaluations of model-generated responses with human judgments. Our work is the first to apply a similar approach to RL, at the same time as some contemporaneous works. \citet{dineen2025qa} use detailed rubrics that greatly improve model safety. Their work differs from ours by using RL with explicit rewards (GRPO) and using checklists that were defined at a global level (the same large set of criteria applied to all tasks) rather than the instruction-specific checklists we espouse in this work.

\section{Limitations}
\label{sec:limitations}
We highlight three limitations with our work. First, our implementation of RLCF uses ``strong-to-weak generalization'' -- a larger model (\texttt{Qwen2.5-72B-Instruct}) provides AI judgments for tuning a smaller model, though RLCF beats other methods also useing a 72B teacher. Second, we only explored preference-based RL in our work. We believe that using checklist feedback to train policy gradient-based algorithms is an exciting future research direction. Lastly, our scoring method is expensive -- grading response pairs on each requirement for 130k instructions with \texttt{Qwen2.5-72B-Instruct} takes roughly 4 days on eight H100 GPUs with 80GB GPU memory, which is computationally infeasible for many practitioners. In \autoref{sec:num_judges}, we show that this cost can be reduced by 50\% at some slight cost to accuracy, but further optimization is warranted.

\section{Conclusion}
We provide a detailed study of reinforcement learning from checklist feedback (RLCF). We propose a novel algorithm for automatically extracting rubrics from instructions, and we use this algorithm to construct a dataset of instructions and rubrics, \textit{WildChecklists}. We demonstrate that RLCF is uniformly effective at improving strong instruction following models on all benchmarks we consider.

Our study follows an active line of work that highlights the limitations of reward models in supervising reinforcement learning. One exciting future direction to emerge from this work is: how can we combine checklist-style feedback with trainable judges? Our current approach relies on carefully-designed, prompt-based components for rubric generation and response grading under a rubric. Why is this more effective than methods that naturally learn to grade responses from human preference data? We believe that analysis of RLCF can motivate better reward models in the future. 

\section*{Acknowledgements}
We thank Saumya Gandhi, Xiang Yue, Gokul Swamy, Apurva Gandhi, Lintang Sutawika, Jessie Mindel, Qianou Ma, Chenyang Yang, and Xinran Zhao for helpful discussions and Akhila Yerukola for invaluable writing assistance and technical advice. We also thank the CMU Foundation and Language Model (FLAME) Center and Amazon Web Services (AWS) for providing computational resources that were essential in executing this work. Vijay Viswanathan, Graham Neubig, and Tongshuang Wu were supported in part by a grant from Apple. Any views, opinions, findings, and conclusions or recommendations expressed in this material are those of the author(s) and should not be interpreted as reflecting the views, policies, or position, either expressed or implied, of Apple Inc.

\bibliographystyle{unsrtnat}
\bibliography{custom}

% Add appendix A
\appendix
\section{The role of response pair mining}

\begin{figure}[h]
    
  \centering
\includegraphics[clip, width=1\linewidth]{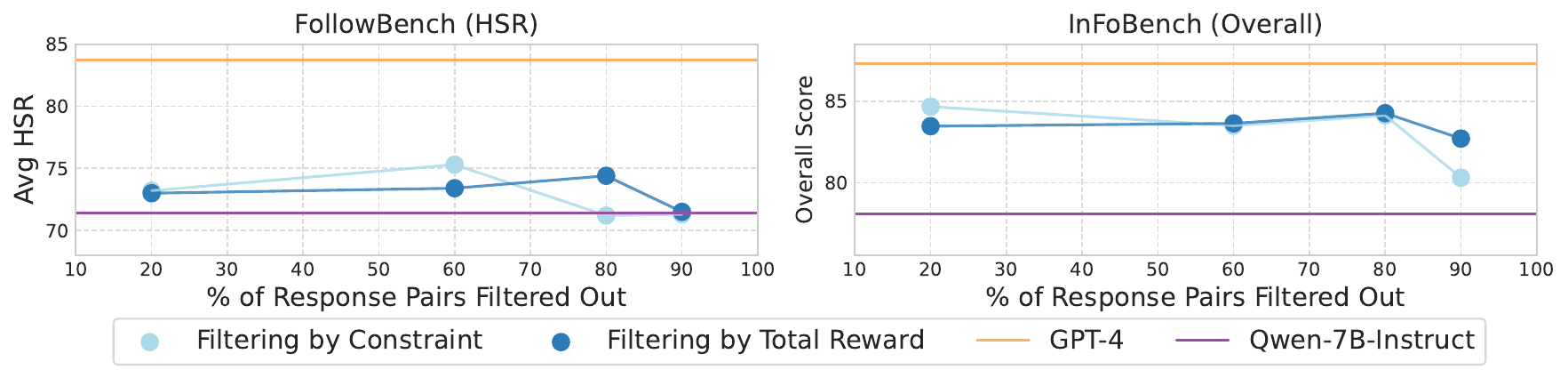}
  \caption{Impact of different filtering strategies on model performance on FollowBench and InFoBench. We compare filtering pairs based on overall checklist score differences versus filtering based on single-aspect score differences, at varying dataset sizes. There are only slight differences between these two filtering methods, until we start filtering out the vast majority of the data. This suggests that the reward signal, rather than the specific filtering algorithm, is likely responsible for this method's effectiveness.}
  \label{fig:filtering_figure}
  \vspace{20pt}
\end{figure}

% \gncomment{I wasn't very sure about the interestingness of the results here with respect to checklist feedback. It's basically saying that filtering out ambiguous examples is a good idea regardless? Maybe this could be moved to an appendix?}
In our algorithm for learning from checklist feedback, we only train on the 40\% of response pairs that differ the most on at least one criterion. This approach differs from thresholding on the reward difference with a single scalar reward, which may represent the aggregation of multiple small differences across all requirements. How much is the filtering component responsible for the success of RLCF?

To investigate this, we compared two approaches: selecting pairs with the largest differences in overall weighted checklist scores versus selecting pairs with the largest differences on any single aspect's score. As shown in Figure \ref{fig:filtering_figure}, performance shows that, when discarding just 20\% or 40\% of response pairs, the method of filtering makes almost no difference. On the other hand, when discarding 90\% of response pairs (with least difference in reward), performance plummets on both benchmarks, suggesting that, regardless of the filtering strategy, keeping some ``harder'' response pairs is beneficial. Rather than aspect-based filtering being the primary driver of improvement, the results suggest that checklist-based rewards inherently capture more instruction-relevant dimensions of quality, leading to more effective preference tuning even with moderate filtering.

% This suggests that while some level of filtering is beneficial, the fundamental quality of the reward signal is more important than extensive filtering. The consistent outperformance of checklist-based rewards over the baseline Instruction-tuned model across all filtering levels indicates that our approach provides a less biased and more generalizable learning signal. Rather than aspect-based filtering being the primary driver of improvement, the results suggest that checklist-based rewards inherently capture more instruction-relevant dimensions of quality, leading to more effective preference tuning even with moderate filtering.

\section{Prompt for Generating Verification Programs}
\label{sec:generate_verification_code}
We describe the prompt used for generating programs to selectively verify responses in \autoref{fig:req_write_code}.

\begin{figure}[t]\centering
\tiny
\begin{minipage}{1.0\columnwidth}\vspace{0mm}    \centering
\begin{tcolorbox} 
    \raggedright
    \small
     \hspace{-6mm}
    \  \\
You are responsible for helping me verify whether or not responses satisfy various requirements. Given a natural language requirement, you will have to classify whether this can be converted to a Python program to automatically check it or whether it should be given to a human collaborator. Your human collaborator is a reliable and cheap expert, and you should trust them. Accordingly, only write code for verifying a constraint if you are very confident that this will exactly check the constraint. You should never make ANY approximations when verifying a constraint. If you feel that you must approximate the constraint in order to verify whether a response follows that constraint, let your human collaborator take care of it. You should ONLY generate code for requirements that are explicitly about syntax or format (e.g. punctuation, unicode characters used, number of paragraphs, shallow grammar, presence of some mandatory keyword specified by the prompt, etc). If there are many different ways to write an answer, you most likely should not generate code for it. If you are not sure, you should not generate code. You should only generate code if you are 100\% sure that the constraint can be verified perfectly with a simple Python function.\\
\vspace{10pt}
When a constraint can be verified EXACTLY with a program, then return a Python function that verifies the constraint. This code should be contained within two sets of triple backquotes, ```. The Python function must return a boolean, and it should only use builtins/standard libraries in Python. If the constraint cannot be verified with a simple Python function (which means your human collaborator will handle the verification of this constraint), please return "NONE" and nothing else. The safest thing to do is to return "defer to human expert \#\#\#\#" 95\% of the time. Now, let's go through a couple examples:\\
\vspace{10pt}
Input:\\
Outline a curriculum development process for a 16-week high school history course, including setting week-by-week objectives and designing assignments. Include two mid-term exams and a final exam. Provide a detailed grading criteria based on the assignments and exams you have designed.\\
\vspace{10pt}
Requirement:\\
Does the response specify the inclusion of two mid-term exams and a final exam\\
\vspace{10pt}
Verification Function:\\
defer to human expert \#\#\#\#\\
(there are multiple valid ways to describe this, and it is not a simple boolean check)\\
...\\
\vspace{10pt}
Input:\\
Welcome to ISLAM STORE's Brand Story\\
Our Journey: A Vision Brought to Life ISLAM STORE was founded with the vision to create an inclusive, informative, and accessible platform for Muslims and non-Muslims alike. Our goal is to promote awareness and understanding of Islam while offering high-quality Islamic products.\\
\vspace{10pt}
Requirement:\\
Does the generated text contain any Arabic?\\
\vspace{10pt}
Verification Function:\\
```python\\
def verify\_requirement(text):\\
    \# Arabic Unicode block range (0600-06FF)\\
    \# Plus Extended Arabic (0750-077F)\\
    \# Plus Arabic Presentation Forms (FB50-FDFF, FE70-FEFF)\\
    return any(('\textbackslash \u0600' <= char <= '\textbackslash \u06FF') or               ('\textbackslash u0750' <= char <= '\textbackslash u077F') or ('\textbackslash uFB50' <= char <= '\textbackslash uFDFF') or            ('\textbackslash uFE70' <= char <= '\textbackslash uFEFF') for char in text)\\
```\\
...\\
\vspace{10pt}
Input:\\
\{input\}\\
\vspace{10pt}
Requirement:\\
\{requirement\}\\
\vspace{10pt}
Verification Function:
\end{tcolorbox}

% \vspace{-2mm}
\caption{
\label{fig:req_write_code}
Prompt for generating verification code}
\end{minipage}
\end{figure}

\section{Prompt for Scoring Semantic Criteria}
\label{sec:scoring_prompt}

\begin{figure}[t]\centering
\tiny
\begin{minipage}{1.0\columnwidth}\vspace{0mm}    \centering
\begin{tcolorbox} 
    \raggedright
    \small
     \hspace{-6mm}
    \  \\
Based on the provided input instruction and response from a worker, assess the response based on the following criteria:\\
1. Does it satisfy the specific requests of the instruction?\\
2. Does the response directly address the request without excessive or off-topic information not necessary for addressing the user's instruction?\\
3. Does the response match the context and the instruction, whether it requires professionalism, friendliness, formality, or neutrality?\\
\vspace{10pt}
Accordingly, score the response with a rating (a number between 0 and 100) assessing how well the response addresses the instruction. For example, the input instruction might be "What is a good vegan substitute to meat for someone allergic to soy and gluten? Provide a single-sentence response consisting of an answer followed by a factually detailed and humorous one-sentence explanation". Your selection should be based on the response and the instruction, using the following rating scale:\\
\vspace{10pt}
- 100: Select 100 if the generated text represents an optimal solution that expertly balances all relevant aspects of the instruction. For the example above (about the vegan substitute), and the criterion above (about factual detail), an example 100-point response is "Mushrooms, because they can be easily caramelized and browned, they are rich in the glutamates which lead to incredible umami flavors, they naturally are completely free of soy and gluten, and they don't look cute as babies". This response is richly detailed and factual, and though it fails to be humorous, it is still a 100-point response on the factual detail criterion.\\
- 75: Return ~75 if the generated text very effectively addresses the main requirements but has room for minor improvements. The response should be unconditionally acceptable (at a professional level) but may not be absolutely perfect. There are no mistakes that critically undermine the question. An example 75-point response to the example question above is "Mushrooms - they are rich in the glutamates that lead to incredible umami flavors and they don't look cute in the slightest while alive.". This response has one interesting fact but could be more detailed.\\
- 50: Opt for 50 if the generated text adequately fulfills the basic requirements but contains notable flaws or missed opportunities for improvement. The response should still be functionally acceptable. The response contains at most one minor inadequacy or inaccuracy related to the question but there are no mistakes that critically undermine the question. An example 50-point response to the example question above is "Mushrooms, because they can be easily caramelized and browned, they're universally beloved by sophisticated palates, and they don't look cute in the slightest while alive." The statement that they're universally beloved by people with sophisticated palates, while potentially true, is vague and not objective.\\
- 25: Return ~25 if the generated text fulfills the key condition specified by the question and demonstrates awareness of the key requirements but fails to execute them effectively. The text may contain non-critical inaccuracies or irrelevant information. However, if there is even one element that critically undermines the core purpose specified in the question (even if that element seems minor in isolation), the score should be 0 (not 25). An example 25-point response to the example question above is "Mushrooms, because they can be easily caramelized and browned, they are absolutely brimming with protein, and they don't look cute in the slightest while alive." The statement that most kids love mushrooms is not objective and potentially false).\\
- 0: Opt for 0 if the generated text fails to meet the question’s requirements or provides no information that could be utilized to answer the question. If the response contains a critical error relevant to the question, return a 0. For the question about the vegan substitute, an example 0-point response is "Mushrooms, because they make you question why you ever thought a dead animal could compare to this vegan delight." While funny and engaging, this response contains zero factual detail about mushrooms, critically violating the question.\\
\vspace{10pt}

Your score can be any number between 0 and 100 (not just the ones listed above). If you are totally confused, return -1 as a default. You should use your judgment to determine the most appropriate score. Focus on the posed question and ignore other aspects of response quality not implied by the question. Return only a number - do not include any other text in your response.\\
\vspace{10pt}

Input:\\
\{instruction\}\\

Generated Text:\\
\{response\}\\

Question:\\
\{requirement\}\\

Score:  \\
\end{tcolorbox}

% \vspace{-2mm}
\caption{
\label{fig:req_check_prompt}
Prompt for checklist scoring}
\end{minipage}
\end{figure}

We describe the prompt used for requirement checking in \autoref{fig:req_check_prompt}.

\end{document}